\documentclass[runningheads]{llncs}

\usepackage[mobile]{eccv}

\usepackage{eccvabbrv}
\usepackage{bm}
\usepackage{wrapfig}

\usepackage{graphicx}
\usepackage{booktabs}

\usepackage{algorithm}
\usepackage[noend]{algpseudocode}

\usepackage{listings}
\usepackage{xcolor}

\usepackage[accsupp]{axessibility}  % Improves PDF readability for those with disabilities.

\usepackage{hyperref}

\usepackage{orcidlink}
\definecolor{Cardinal}{rgb}{0.549,0.082,0.082}

\newcommand{\myparagraph}[1]{\vspace{0.15cm}\noindent\emph{#1}}

\newcommand{\model}{PhysDreamer\xspace}

\begin{document}

\newcommand{\bx}{\boldsymbol{x}}
\newcommand{\bv}{\boldsymbol{v}}
\newcommand{\bu}{\boldsymbol{u}}

\newcommand{\bsigma}{\boldsymbol{\sigma}}

\newcommand{\bF}{\boldsymbol{F}}
\newcommand{\bR}{\boldsymbol{R}}
\newcommand{\bC}{\boldsymbol{C}}

% \title{Physics-based 3D Object Motion Synthesis \\via Video Generation} 

% \title{i-Gaussian: \\Physics-based Interaction with 3D Gaussians} 

\title{PhysDreamer: Physics-Based Interaction with \\ 3D Objects via Video Generation}

%\title{Poking a Rose in Five Thousand Ways}

\titlerunning{PhysDreamer}

% Second Author\inst{2,3}\orcidlink{1111-2222-3333-4444} \and
% Third Author\inst{3}\orcidlink{2222--3333-4444-5555}}

\author{Tianyuan Zhang\inst{1}  \and
Hong-Xing Yu \inst{2} \and
Rundi Wu \inst{3} \and 
Brandon Y. Feng \inst{1} \and \\
Changxi Zheng \inst{3} \and 
Noah Snavely \inst{4} \and 
Jiajun Wu \inst{2} \and 
William T. Freeman \inst{1}
}

\authorrunning{Zhang et al.}

\institute{Massachusetts Institute of Technology \and
Stanford University \and Columbia University \and Cornell University
}

\maketitle

\begin{abstract}
  % Given an elastic object represented as 3D Gaussians, we present an approach to synthesize realistic motions of it under physical interaction, e.g., its motion after being poked. We model the object's dynamics with hyper-elastic materials and simulate it with material point methods. We find a set of spatially varying material parameters by distilling knowledge of object's dynamics from video generative models. With distilled material parameters, we make the static 3D objects interactable, enabling us not only to render it photo-realistic, but also to simulate its motions realistically under new interactions.  
  
  % We present \model, an approach to synthesizing realistic, physics-based 3D object dynamics under interaction. Our method estimates an object's complex, spatially varying physical material properties from its static 3D representation by leveraging image-to-video generation and differentiable simulation. We optimize the material properties so that a rendered video of the simulated 3D dynamics aligns with a generated reference video that depicts the object in plausible motion. \model enables the synthesis of realistic dynamics of 3D objects under novel interactions, such as external forces or agent manipulations. We demonstrate our approach with diverse examples of elastic objects and evaluate the realism of the synthesized motion through a user study. Our work endows the dynamic interactivity of static 3D objects and enhances their realism, taking a step towards more immersive virtual experiences and realistic simulations.

  Realistic object interactions are crucial for creating immersive virtual experiences, yet synthesizing realistic 3D object dynamics in response to novel interactions remains a significant challenge. Unlike unconditional or text-conditioned dynamics generation, action-conditioned dynamics requires perceiving the physical material properties of objects and grounding the 3D motion prediction on these properties, such as object stiffness. However, estimating physical material properties is an open problem due to the lack of material ground-truth data, as measuring these properties for real objects is highly difficult. We present \model, a physics-based approach that endows static 3D objects with interactive dynamics by leveraging the object dynamics priors learned by video generation models. By distilling these priors, \model enables the synthesis of realistic object responses to novel interactions, such as external forces or agent manipulations. We demonstrate our approach on diverse examples of elastic objects and evaluate the realism of the synthesized interactions through a user study. \model takes a step towards more engaging and realistic virtual experiences by enabling static 3D objects to dynamically respond to interactive stimuli in a physically plausible manner. See our project page at \url{https://physdreamer.github.io/}.

\end{abstract}

\section{Introduction}
\label{sec:intro}

Realistic object interactions play a pivotal role in creating immersive virtual experiences. Recent advances in 3D vision have enabled the capture and creation of high-quality static 3D assets~\cite{mildenhall2020nerf,gaussiansplat2023}, and some methods even extend to 4D assets~\cite{dynamicgaussluiten2023,dreamgaussian4d2023,alignling2023}, generating unconditioned dynamics. However, these methods fail to handle action-conditioned dynamics in response to new physical interactions, such as synthesizing the motion of a rose reacting to a breeze or a touch.

% which is crucial for synthesizing realistic object responses to novel interactions, such as external forces or agent manipulations.

% Action-conditioned dynamics differs from unconditional and text-conditioned dynamics in that it requires perceiving the physical material properties of objects and grounding the 3D motion prediction on these properties, such as object stiffness. Yet, estimating physical material properties is a challenging task due to the lack of ground-truth data, as measuring these properties for real objects is highly difficult. Real-life objects often exhibit complex, spatially-varying material properties, making the estimation problem even more challenging.

The key challenge in synthesizing action-conditioned dynamics lies in understanding the physical material properties of objects. Yet, estimating these properties is a challenging task due to the lack of ground-truth data, as measuring these properties for real objects is highly difficult. Real-life objects often exhibit complex, spatially-varying material properties, making the estimation problem even more challenging. Despite the complexity of physical materials, humans can easily imagine how objects would react to external forces, such as the gentle sway of a rose. This ability to imagine object dynamics stems from our physical prior knowledge obtained from observing and interacting with the physical world. This motivates us to distill dynamics priors from video generation models that have been trained on vast, diverse video observations of the physical world. 
% harness and ground the physical priors learned in video generation models. 

\begin{figure}[t]
    \centering
    \includegraphics[width=\columnwidth]{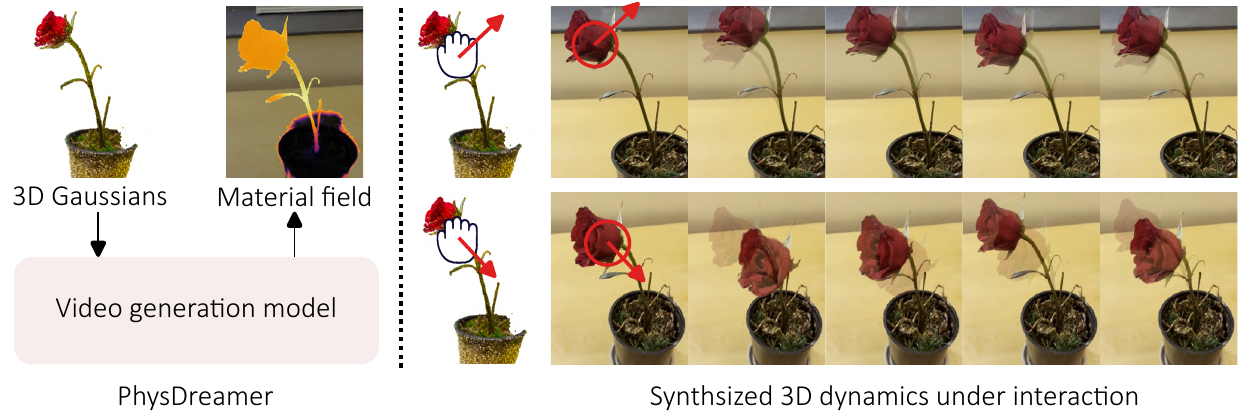}
    % \put(-325,0){\model}
    % \put(-180,0){Synthesized 3D motion under interactions}
    \vspace{-0.4cm}
    \caption{ \textbf{(Left)} Leveraging and distilling dynamics priors from a pre-trained video generation model, we estimate a physical material field for the static 3D object. \textbf{(Right)} The physical material field allows synthesizing interactive 3D dynamics under arbitrary forces. We show rendered sequences from two viewpoints. Red arrows indicate force directions. Please see videos on our project website for better visualization.
    }
    \label{fig:teaser}
    \vspace{-8pt}
\end{figure}

In this work, we focus on synthesizing interactive 3D dynamics. We propose \textbf{\model}, a physics-based approach to transforming static 3D objects into interactive ones that can respond to novel interactions. 
% To address this challenge, we propose \textbf{\model}, a physics-based approach that leverages the object dynamics priors learned by video generation models. 
The key idea behind \model is to distill dynamics priors learned by video generation models to estimate the physical material properties of static 3D objects. We hypothesize that video generation models, trained on large amounts of video data, implicitly capture the relationship between object appearance and dynamics. By leveraging this learned prior knowledge, \model can infer the physical material properties that drive the dynamic behavior of objects, even in the absence of ground-truth material data (Fig.~\ref{fig:teaser}).

\model represents 3D objects using 3D Gaussians~\cite{gaussiansplat2023}, models the physical material field with a neural field~\cite{xie2022neural}, and simulates 3D dynamics using the differentiable Material Point Method (MPM)~\cite{materialjiang2016, physgaussianxie2023}. The differentiable simulation and rendering allow for direct optimization of the physical material field and initial velocity field by matching pixel space observations. We focus on elastic dynamics and showcase \model through diverse real examples, such as flowers, plants, a beanie hat, and a telephone cord. We evaluate the realism of the synthesized interactive motion through a user study, comparing \model to state-of-the-art methods. The results demonstrate that our approach significantly outperforms existing techniques on motion realism and visual quality. 
% validating the effectiveness of leveraging video generation priors for estimating physical material properties and synthesizing interactive 3D dynamics.

In summary, \model addresses the challenge of synthesizing interactive 3D dynamics by leveraging the object dynamics priors learned by video generation models. By distilling these priors to estimate the physical material properties of static 3D objects, our approach enables the creation of immersive virtual experiences where objects can respond realistically to novel interactions. The main contributions of our work include enabling static 3D objects to dynamically respond to interactive stimuli in a physically plausible manner and taking a step towards more engaging and realistic virtual experiences. We believe that \model has the potential to greatly enhance the realism and interactivity of virtual environments, paving the way for more engaging and lifelike simulations.

\section{Related work}
\label{sec:related}

\subsection{Dynamic 3D reconstruction}

Dynamic 3D reconstruction methods aim to reconstruct a representation of a dynamic scene from inputs such as depth scans\cite{volumetriccurless1996, globalli2008},  RGBD videos\cite{dynamicfusionnewcombe2015}, or monocular or multi-view videos \cite{spacexian2021,nerfiespark2021, dnerfpumarola2021, hypernerfpark2021,dylinyu2023, hyperreelattal2023,  dynamicgaussluiten2023, dynmfkratimenos2023, sax_nerf, cogsyu2023, dynibarli2023, flowwang2023}. 
This task is especially challenging in the monocular setting with slow-moving cameras and fast-moving scenes \cite{monoculargao2022}. 
Novel scene representations are a major driver of recent progress. 
One prominent approach is to augment a canonical Neural Radiance Fields (NeRF) with a deformation field \cite{dnerfpumarola2021}. 
This approach can be further improved by incorporating flow supervision \cite{flowwang2023, forwardflowguo2023} or as-rigid-as-possible or volume preserving regularization terms \cite{nerfiespark2021, hypernerfpark2021}. 
Time-modulated NeRFs \cite{nsffli2021, monoculargao2022, kplanefridovich2023, hexplanecao2023} offer a simpler alternative representation. 
Due to its Lagrangian nature, 3D Gaussian Splatting\cite{gaussiansplat2023} is readily adaptable to the task of efficient dynamic scene reconstruction \cite{dynamicgaussluiten2023, dynmfkratimenos2023, cogsyu2023, deformablegaussianyang2023, scgshuang2023, 4dgsduan2024}.
Data-driven prior, such as from monocular depth models \cite{consistentzhang2021, robustcvdkopf2021} and image diffusion models \cite{diffusionwang2024}, can also be used to reduce the inherent ambiguity in dynamic reconstruction from monocular videos.

\subsection{Dynamic 3D generation}

Our work also relates to efforts to synthesize dynamic 3D scenes. 
A common approach is to integrate a 3D generation pipeline with a video generation model \cite{text4dsinger2023, 4dfybahmani2023, alignling2023, dreamgaussian4d2023}. 
For instance, Make-A-Video3D begins by creating a static NeRF as per DreamFusion \cite{dreamfusionpoole2022}, then extending it temporally using Score Distillation Sampling (SDS) \cite{dreamfusionpoole2022} derived from a video diffusion model. 
The approach can be improved with more efficient representations, stronger diffusion priors, and stable training techniques \cite{4dfybahmani2023, alignling2023}. 
However, applying SDS with video diffusion models demands significant computational and memory costs. 
Compact4D \cite{4dgenyin2023} and DreamGaussian4D \cite{dreamgaussian4d2023} used a more efficient approach, synthesizing 3D dynamics by aligning a reference video from video generation models while employing SDS from image diffusion models to reduce novel view artifacts. These methods are currently limited to producing fixed-length 3D videos. We focus on synthesizing interactive 3D motions under any new physical interactions.

\subsection{Interactive motion generation}

Interactive motion generation animates still images or 3D contents according to user inputs like text \cite{livephotochen2023, i2vgenzhang2023}, motion fields \cite{motionguidancegeng2023}, motion layers \cite{animateanythingdai2023, stochastic_textures_chuang2005}, or direct manipulation such as dragging and pulling \cite{imagemodaldavis2015, generativeli2023}.  
Early work from Davis et al.~\cite{imagemodaldavis2015, visualvibrationdavis2016} demonstrated animating an image using an image-space modal basis extracted from a video of an object undergoing subtle vibrational motions. 
Building upon this image-space representation\cite{imagemodaldavis2015}, Generative Image Dynamics \cite{generativeli2023} used a diffusion model trained on a dataset with paired image and its modal basis to model  scene motion distributions, enabling realistic interaction with still input images. We focus on interacting with 3D objects rather than images. 

For 3D assets, physics-based approaches enable synthesizing motions under any physical interactions. 
Virtual Elastic Objects\cite{virtual_elastic_objects_chen2022} jointly reconstructs the geometry, appearances, and physical parameters of elastic objects in a multiview capture setup with compressed air system.
PAC-NeRF \cite{pacli2023}, DANO \cite{le2023differentiable}, and PhysGaussian \cite{pienerffeng2023} integrate physics-based simulations with NeRF and 3D Gaussians to generate physically plausible motions. We use the same physics-based approach to generate realistic interactions, but a novel ingredient of our work is to distill the material parameters of the object from pre-trained video generation models.

\subsection{Video generation models}

Recent progress in video generation is driven by the development of larger autoregressive \cite{phenakivillegas2022, cogvideohong2022, nuwawu2022, videopoetkondratyuk2023} and diffusion models \cite{makeavideo2022, imagenvideo2022, alignyourlatents2023, emuvideo2023, svdblattmann2023, lumiere2024, sora2024, waltgupta2023}.  
These models, trained on increasingly large datasets, continue to advance the quality and realism of generated video content.  The state-of-the-art approach \cite{sora2024} can generate minute-long videos with realistic motions and viewpoint consistency. 
However, current video generation models cannot support physics-based interactions with objects through external forces. 
% Our approach leverages strong motion prior from a video generation model, and as the field of video generation continues to advance, we expect our method to yield even better results.
% We ground the neural motion prior in pixel space to/with? a physics-based model, enabling interactive synthesis from physical interaction.

% Common techniques to improve training efficiency include ``inflating'' pretrained text-to-image models \cite{alignyourlatents2023, emuvideo2023,  svdblattmann2023}, successive temporal and spatial downsampling \cite{lumiere2024}, improved tokenizers  \cite{waltgupta2023, videopoetkondratyuk2023} and more scalable architectures  \cite{videopoetkondratyuk2023, sora2024}. 

% \input{sections/03_background}

% \input{sections/04_method_new}

\section{Problem formulation}

\begin{figure}[t]
    \centering
    \includegraphics[width=1.0\columnwidth]{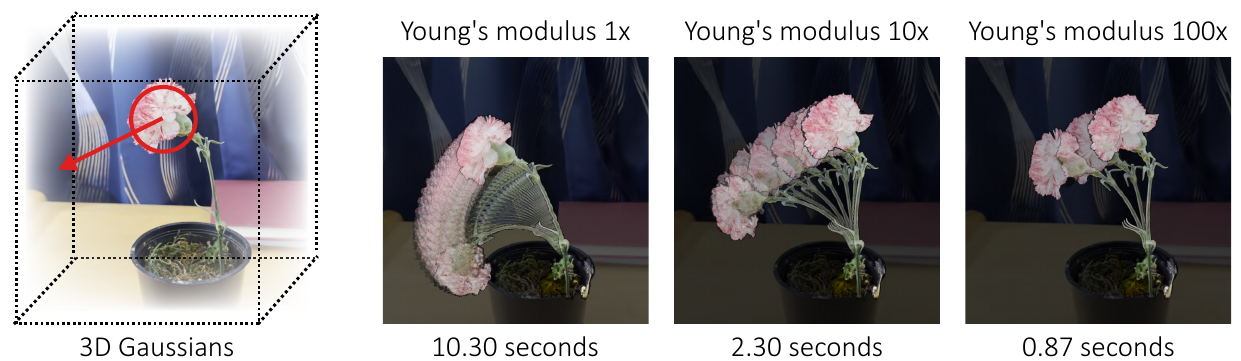}
    % \put(-50,1){How Young's modulus affect the motion}
    \caption{Effect of Young's modulus. We depict the motion of a simulated flower under the same external force but with three different Young's moduli, a measure of material stiffness. Flowers with the highest Young's modulus (100$\times$) exhibit smaller oscillations and higher frequencies, while the flower with the lowest Young's modulus (1$\times$) sways the most and oscillates at the lowest frequency. Time annotations below each image indicate the duration of one complete motion path shown in the figure.}
    \label{fig:young}
    \vspace{-8pt}
\end{figure}

Given a static object represented by 3D Gaussians $\{\mathcal{G}_p\}_{p=1}^P$, $\mathcal{G}_p = \{\bx_p, \alpha_p, \bm{\Sigma}_p, \bm{c}_p \}$ (where $\bx_p$ denotes the position, $\alpha_p$ denotes the opacity, $\bm{\Sigma}_p$ denotes the covariance matrix, and $\bm{c}_p$ denotes the color of the particle), our goal is to estimate physical material property fields for the object to enable realistic interactive motion synthesis. These properties include mass $m$, Young's modulus $E$, and Poisson's ratio $\nu$.
Among these physical properties, Young's modulus $E$ plays a particularly important role in determining the object's motion in response to applied forces.
Intuitively, Young's modulus (Eq.~\ref{eq:young_poisson}) measures the material stiffness. 
A higher Young's modulus results in less deformation and more rigid and higher-frequency motion, while a lower value leads to more flexible and elastic behavior. Fig.~\ref{fig:young} illustrates the simulated motion of a flower under the same applied forces but with different Young's modulus.

Therefore, our problem formulation focuses on estimating the spatially varying Young's modulus field $E(\bx)$ for the 3D object. 
To allow particle simulation, we query a particle's Young's modulus by $E_p=E(\bx_p)$.
As for other physical properties, the mass for a particle $m_p$ can be pre-computed as the product of a constant density ($\rho$) and particle volume $V_p$. 
The particle volume can be estimated~\cite{physgaussianxie2023} by dividing a background cell's volume by the number of particles that cell contains. 
As for the Poisson's ratio $\nu_p$, we found that it has negligible impact on object motion in our preliminary experiments(see supplementary materials for details), and so we assume a homogeneous constant Poisson's ratio.

\section{\model}

\begin{figure}[t]
    \centering
    \includegraphics[width=1.0\columnwidth]{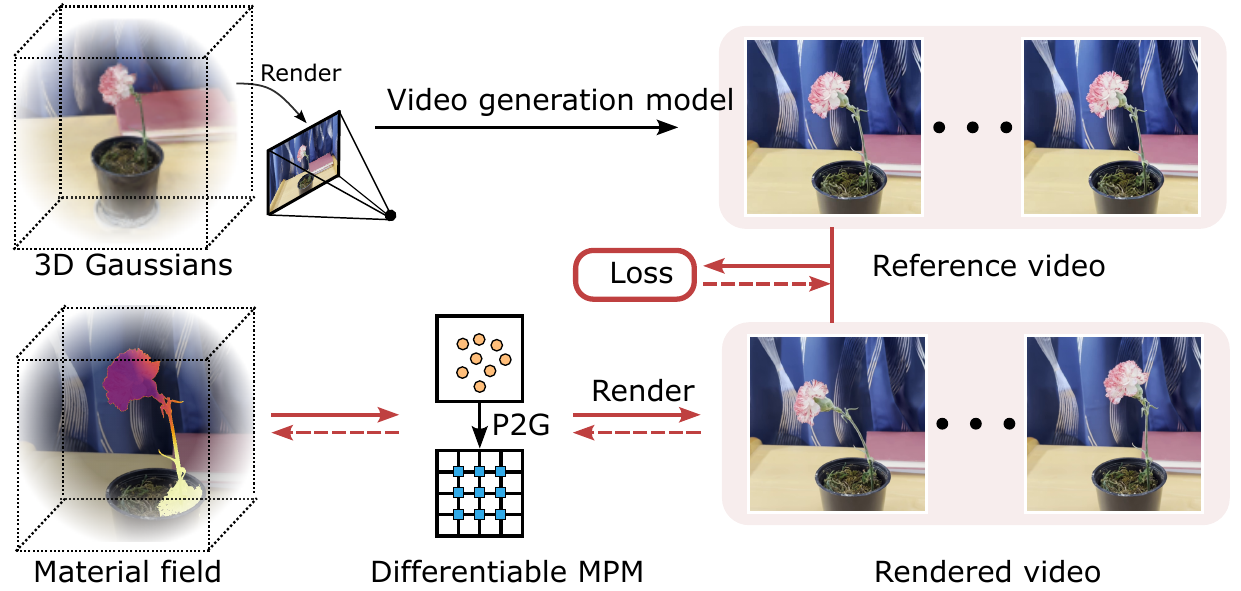}
    \caption{Overview of \model.  Given an object represented as 3D Gaussians, we first render it (with background) from a viewpoint. Next, we use an image-to-video generation model to produce a reference video of that object in motion. Using differentiable Material Point Methods (MPM) and differentiable rendering, we optimize both a spatially-varying material field and an initial velocity field (not shown in the figure above). This optimization aims to minimize the discrepancy between the rendered video and the reference video. The dashed arrows represent gradient flow. }
    \label{fig:method_main}
    \vspace{-8pt}
\end{figure}

\model estimates a material field for a static 3D object. Our key idea is to generate a plausible video of the object in motion, and then optimize the material field $E(\bx)$ to match this synthesized motion.
We begin by rendering a static image ($I_0$) for the 3D scene $\{\mathcal{G}_p\}$ from a certain viewpoint. We then leverage an image-to-video model to generate a short video clip $\{I_0, I_1, \ldots, I_T\}$ depicting the object's realistic motion. This generated video serves as our reference video. We then optimize the material field $E(\bx)$ and an initial velocity field $\bv_0(\bx)$ (both modeled by implicit neural fields~\cite{xie2022neural}) through differentiable simulation and differentiable rendering, such that a rendered video of the simulation matches (from the same viewpoint as $I_0$) the reference video.
Fig.~\ref{fig:method_main} shows an overview of \model.

\subsection{Preliminaries}
3D Gaussians~\cite{gaussiansplat2023} adopts a set of anisotropic 3D Gaussian kernels to represent the radiance field of a 3D scene. 
Although introduced primarily as an efficient method for 3D novel view synthesis, the Lagrangian nature of 3D Gaussians also enables the direct adaptation of particle-based physics simulators. 
Following PhysGaussian~\cite{physgaussianxie2023}, we use the Material Point Method (MPM) to simulate object dynamics directly on these Gaussian particles. 
Since 3D Gaussians mainly lie on object surfaces, an optional internal filling process can be applied for improved simulation realism\cite{physgaussianxie2023}.  
Below, we provide a brief introduction on the underlying physical model and how to integrate MPM into 3D Gaussians. 
For a more comprehensive introduction of MPM, we refer interested readers to \cite{materialjiang2016, apicjiang2015, mlsmpmhu2018, physgaussianxie2023}.

\myparagraph{Continuum mechanics and elastic materials.}
Continuum mechanics models material deformation using a map $\phi$ that transforms points from the undeformed material space $\mathbf{X}$ to the deformed world space $\mathbf{x} = \phi(\mathbf{X}, t)$. 
The Jacobian of the map, $\boldsymbol{F} = \nabla_{\mathbf{X}} \phi(\mathbf{X}, t)$, known as the deformation gradient, measures local rotation and strain.
This tensor is crucial in formulating stress-strain relationship. 
For example, the Cauchy stress in a hyper-elastic material is computed by: $ \bsigma = \frac{1}{\mathrm{det}(\bF)} \frac{\partial \psi}{\partial \bF} \bF^T$. 
Here, $\psi(\mathbf{F})$ represents the strain energy density function, quantifying the extent of non-rigid deformations.
This function is typically designed by experts, to follow principles like material symmetry and rotational invariance while aligning with empirical data. 
In this work, we use fixed corotated hyperelastic model, whose energy density function can be expressed as: 
\begin{equation}
    \label{eq:fixed-corotated}
\psi(\mathbf{F}) =  \mu \left( \sum_{i=1}^{d} (\sigma_i - 1)^2 \right) + \frac{\lambda}{2}(\mathrm{det}(\bF) - 1)^2,
\end{equation}
where $\sigma_i$ denotes a singular value of the deformation gradient. $\mu$ and $\lambda$ are related to Young's modulus $E$ and Poisson's ratio $\nu$ via: 
\begin{equation}
\label{eq:young_poisson}
    \mu = \frac{E}{2(1+\nu)}, \quad \lambda = \frac{E\nu}{(1+\nu)(1-2\nu)}. 
\end{equation}
The dynamics of an elastic object are governed by the following equations: 
\begin{equation}
    \begin{aligned}
        \rho \frac{D \bv }{D t} = \nabla \cdot \bsigma + \mathbf{f},  \quad
        \frac{D\rho}{Dt} + \rho \nabla \cdot \bv = 0,  \\
    \end{aligned}
\end{equation}
where $\rho$ denotes density, $\bv(\bx, t)$ denotes the velocity field in world space, and $\mathbf{f}$ denotes an external force.

\myparagraph{Material Point Method (MPM).}
We use the Material Point Method (MPM)~\cite{materialjiang2016, physgaussianxie2023} to solve the above governing equation. 
MPM is a hybrid Eulerian-Langrangian method widely adopted for simulating dynamics for a wide range of materials, such as solid, fluid, sand, and cloth \cite{materialram2015, druckerklar2016, materialstomakhin2013, jiang2017anisotropic}.
MPM offers several advantages, such as easy GPU parallelization~\cite{hu2019taichi}, handling of topology changes, and the availability of well-documented open-source implementations~\cite{taichihu2019, warp2022, nclawma2023, physgaussianxie2023}.

Following PhysGaussian~\cite{physgaussianxie2023}, we view the Gaussian particles as the spatial discretization of the object to be simulated, and directly run MPM on these Gaussian particles.
Each particle $p$ represents a small volume of the object, and it carries a set of properties including volume $V_p$, mass $m_p$, position $\bx_p^t$, velocity $\bv_p^t$, deformation gradient $\bF_p^t$, and local velocity field gradient $\boldsymbol{C}_p^t$ at time step $t$. 

MPM operates in a particle-to-grid (P2G) and grid-to-particle (G2P) transfer loop. 
In the P2G stage, we transfer the momentum from particle to grid by: 
\begin{equation}
    m_i^t \bv_i^{t} = \sum_p N(\bx_i - \bx_p^t) [m_p \bv_p^t + (m_p\bC_p^t - \frac{4}{(\Delta x)^2} \Delta t V_p  \frac{\partial \psi}{\partial \bF} {\bF_p^t}^T )(\bx_i - \bx_p^t) ] + \boldsymbol{f}_i^t, 
\end{equation}
where the mass of the grid node $i$ is $m_i^t = \sum_p N(\bx_i - \bx_p^t) m_p$, $N(\bx_i - \bx_p^t)$ is the B-spline kernel, $\Delta x$ is the spatial grid resolution, $\Delta t$ is the simulation step size, and $\bv_i^{t}$ is the updated velocity on the grid. We then transfer the updated velocity back to the particles and update their positions as:
\begin{equation}
    \bv_p^{t+1} = \sum_i N(\bx_i - \bx_p^t) \bv_i^{t}, \quad \bx_p^{t+1} = \bx_p^{t} + \Delta t \bv_p^{t+1}. 
\end{equation}
Meanwhile, the local velocity gradient and deformation gradient is updated as: 
% \begin{equation}
%    \bC_p^{t+1} = \frac{4}{(\Delta x)^2} \sum_i N(\bx_i - \bx_p^t) \bv_i^{t} (\bx_i - \bx_p^t)^T,  \quad \bF_p^{t+1} = (\boldsymbol{I} + \Delta t \bC_p^{t+1}) \bF_{p}^{t}. 
% \end{equation}
\begin{equation}
   \bC_p^{t+1}=\frac{4}{(\Delta x)^2} \sum_i N(\bx_i - \bx_p^t) \bv_i^{t} (\bx_i - \bx_p^t)^T,  \bF_p^{t+1} = (\boldsymbol{I} + \Delta t  \sum_i \boldsymbol{v}_i^{t} \nabla N(\bx_i - \bx_p^{t})^T ) \bF_{p}^{t}. 
\end{equation}

\subsection{Estimating physical properties}
Using MPM~\cite{materialjiang2016, physgaussianxie2023} as our physics simulator and the Fixed Corotated hyper-elastic material model for the 3D objects, the simulation process for a single sub-step is formalized as: 
\begin{equation}
    \bx^{t+1}, \bv^{t+1}, \bF^{t+1}, \bC^{t+1} = \mathcal{S}(\bx^{t}, \bv^{t}, \bF^{t}, \bC^{t}, \bm{\theta}, \Delta t),
\end{equation}
where $\bx^{t} = [\bx^t_1, \cdots, \bx_P^t]$ denotes the positions of all particles at time $t$, and similarly $\bv^{t} = [\bv^t_1, \cdots, \bv_P^t]$ denotes the velocities of all particles at time $t$. $\bF^{t}$ and $\bC^{t}$ denote the deformation gradient and the gradient of local velocity fields for all particles, respectively. Both $\bF^{t}$ and $\bC^{t}$ are tracked for simulation purposes, not for rendering. $\bm{\theta}$ denotes the collection of the physical properties of all particles: mass $\bm{m} = [m_1, \cdots, m_P]$, Young's modulus $\bm{E} =[E_1, \cdots, E_P]$,  Poisson's ratio $\bm{\nu}=[\nu_1, \cdots, \nu_P]$, and volume $\bm{V}=[V_1, \cdots, V_P]$. $\Delta t$ is the simulation step size. 

We use a sub-step size $\Delta t \approxeq
 1 \times 10^{-4}$ for most of our experiments. To simulate dynamics between adjacent video frames, we iterate over hundreds of sub-steps (time interval between frames are typically tens of milliseconds).  For simplicity, we abuse notation to express a simulation step with $N$ sub-steps as:
\begin{equation}
    \begin{aligned}
    \bx^{t+1}, \bv^{t+1}, \bF^{t+1}, \bC^{t+1} = \mathcal{S}(\bx^{t}, \bv^{t}, \bF^{t}, \bC^{t}, \bm{\theta}, \Delta t, N) , \\
    \end{aligned}
\end{equation}
where the timestamp $t+1$ is ahead of timestamp $t$ by $N \Delta t$. 
After simulation, we render the Gaussians at each frame:
\begin{equation}
    \hat{I}^{t} = \mathcal{F}_{\mathrm{render}} (\bx^{t}, \bm{\alpha}, \bm{R}^{t}, \Sigma, \bm{c}),
\end{equation}
where $\mathcal{F}_{\mathrm{render}}$ denotes the differentiable rendering function, and $\bm{R}^{t}$ denotes the rotation matrices of all particles obtained from the simulation step.  

Using the generated video as reference, we optimize the spatially-varying Young's modulus $\bm{E}$ and an initial velocity $\bv^0$ by a per-frame loss function:
\begin{equation}
\label{eq:rendering_loss}
L^t = \lambda L_{\mathrm{1}} (\hat{I}^t, I^t) + (1 - \lambda) L_{\mathrm{D-SSIM}} ( \hat{I}^t, I^t),
\end{equation}
where we set $\lambda=0.1$ in our experiments. 

We parameterize the material field and velocity field by two triplanes\cite{chan2022efficient}, each followed by a three-layer MLP. Additionally, we apply a total variation regularization for all spatial planes of both fields to encourage spatial smoothness. Using $\bu$ to denote one of the 2D spatial planes, and $\bu_{i,j}$ as  a feature vector on the 2D plane, we write the total variation regularization term as: 
\begin{equation}
    \label{eq:tv_regularization}
    L_{\text{tv}} = \sum_{i,j} \|\bu_{i+1, j} - \bu_{i,j}\|_2^2 + \|\bu_{i, j+1} - \bu_{i,j}\|_2^2.
\end{equation} 

Rather than optimizing the material parameters and initial velocity jointly, we split the optimization into two stages for better stability and faster convergence. 
In particular, in the first stage, we randomly initialize the Young's modulus for each Gaussian particle and freeze it. 
We optimize the initial velocity of each particle using only the first three frames of the reference video.  
In the second stage, we freeze the initial velocity and optimize the spatially varying Young's modulus. 
During the second stage, the gradient signal only flows to the previous frame to prevent gradient explosion/vanishing.

\subsection{Accelerating simulation with subsampling}

\begin{figure}[!t]
\begin{minipage}{0.42\textwidth}
     \captionof{figure}{Accelerated MPM with K-Means downsampling. We employ K-Means clustering to create a set of ``driving particles'' (in yellow) at the initial time step (t=0). We only simulate these driving particles. When rendering, we obtain each particle's position and rotation by fitting a local rigid body transformation using neighboring driving particles.}
    \label{fig:subsample}
\end{minipage}%
\hspace{0.06\textwidth}
\begin{minipage}{0.5\textwidth}
\includegraphics[width=\linewidth]{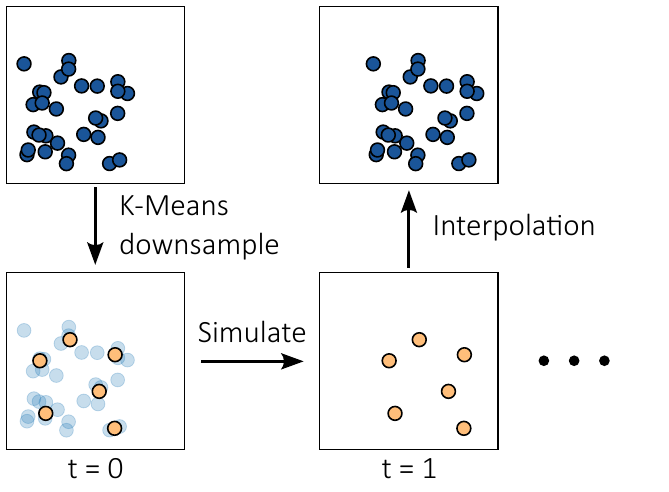}   
\end{minipage}
\vspace{-0.4cm}
\end{figure}

High-fidelity rendering with 3D Gaussians typically requires millions of particles to represent a scene. Running simulations on all the particles poses a significant computational burden. To improve efficiency, we introduce a subsampling procedure for simulation, as illustrated in Fig.~\ref{fig:subsample}.

Specifically, we apply K-Means clustering to create a set of driving particles $\{\mathcal{Q}_q\}_{q=1}^Q$ at $t=0$, where each driving particle is represented by $\mathcal{Q}^0_q = \{\bx^0_q, \bv^0_q, \bF^0_q, \bC^0_q, E_q, m_q, \nu_q, V_q\}$. 
The initial position of a driving particle $\bx_q^0$ is computed as the mean of the position $\bx_p$ of all cluster members. 
The number of the driving particles is much smaller than the number of 3D Gaussian particles, $Q \ll P$. 
We run simulations only on the driving particles. 
During rendering, we compute the position and rotation for each 3D Gaussian particle $\mathcal{G}_p$ by interpolating the driving particles. 
In particular, for each 3D Gaussian particle, we find its eight nearest driving particles at $t=0$, and we fit a rigid body transformation $\bm{T}$ between these eight driving particles at $t=0$ and at the current timestamp. This rigid body transformation $\bm{T}$ is applied to the initial position and rotation of the particle $\mathcal{G}_p$ to obtain its current position and rotation. We summarize our algorithm with pseudo-code in supplementary materials. %Appendix~\ref{sec:algorithm}.

% \begin{wrapfigure}{r}{0.5\textwidth}
%     \vspace{-0.6cm}
%     \includegraphics[width=0.95\linewidth]{figures/figure_acclerate.pdf}
%     \caption{Acclerate MPM with downsampling. We employ K-means clustering to create a set of representative "driving particles" (shown in yellow) at the initial timestep (t=0). We only simulate the drive points. When rendering, we interpolate all the points by fitting a local rigid body transformation.}
%     \vspace{-0.5cm}
%     \label{fig:subsample}
%     \vspace{-0.2cm}
% \end{wrapfigure}

% try to move caption to the right hand side. 

\section{Experiments}
\label{sec:experiment}

\subsection{Setup}

\myparagraph{Datasets.}
We collect eight real-world static scenes by capturing multi-view images.
Each scene includes an object and a background. The objects include five flowers (a red rose, a carnation, an orange rose, a tulip, and a white rose), an alocasia plant, a telephone cord, and a beanie hat. For each scene except for the red rose scene, we capture four interaction videos illustrating its natural motion after interaction, such as poking or dragging, and we use the real videos as additional comparison references.

\myparagraph{Baselines.} 
We compare our approach to two baselines: PhysGaussian~\cite{physgaussianxie2023} and DreamGaussian4D~\cite{dreamgaussian4d2023}. PhysGaussian~\cite{physgaussianxie2023} integrates MPM simulation to static 3D Gaussians to support simulation, but it cannot estimate material properties and relies on manually setting material parameter values. Thus, we use the same initialization strategy as ours to assign material properties for PhysGaussian.
DreamGaussian4D~\cite{dreamgaussian4d2023} generates non-interactive dynamic 3D Gaussians from a static image. It first obtains a static 3D Gaussians using DreamGaussian~\cite{tang2023dreamgaussian}, and then animate it by optimizing a deformation field from a generated driving video.
% by animating generated 3D Gaussians with a generated video using Score Distilliation Sampling (SDS) with a multiview diffusion model and an MSE loss.
For a fair comparison, we run its deformation field optimization on our reconstructed static 3D Gaussians, and we looped the resulting deformation field when rendering longer videos in later comparison.

\myparagraph{Evaluation metrics.}
We focus on the quality of the synthesized object motion, in particular, \emph{visual quality} and \emph{motion realism}. Therefore, we conduct a user study and adopt the Two-alternative Forced Choice (2AFC) protocol: the participants are shown two side-by-side synchronized videos, including one video result from ours and the other one from the competitor’s, with a random left-right ordering. 
The participants are then asked to choose the one with higher visual quality and the one with higher motion realism. 

We recruited $100$ participants, each asked to judge all $8$ scenes, forming a total of $800$ 2AFC judgement samples for each baseline comparison. For each scene, we create $4$ sample video pairs and show participants a random one from the $4$ pairs. In particular, we create $4$ five-second motion sequences using \model with randomized initial conditions (applying an external force to the foreground object or assigning an initial velocity to the object), and render videos from randomly picked viewpoints.
For the baseline method, we apply the same initial conditions (for PhysGaussian only) and render videos from the same viewpoint as ours to form the video pairs.
Please see supplementary materials for human study details and quantitative metrics for videos (e.g., Fréchet Video Distance\cite{fvdunterthiner2018}).
% The participants are asked the following question:
% \emph{``Compare the two moving objects below. Which one has a \textbf{higher visual quality}? That is, which looks better to you and has fewer errors?''} for visual quality and 

\begin{figure}[t]
    \centering
    \includegraphics[width=1.0\columnwidth]{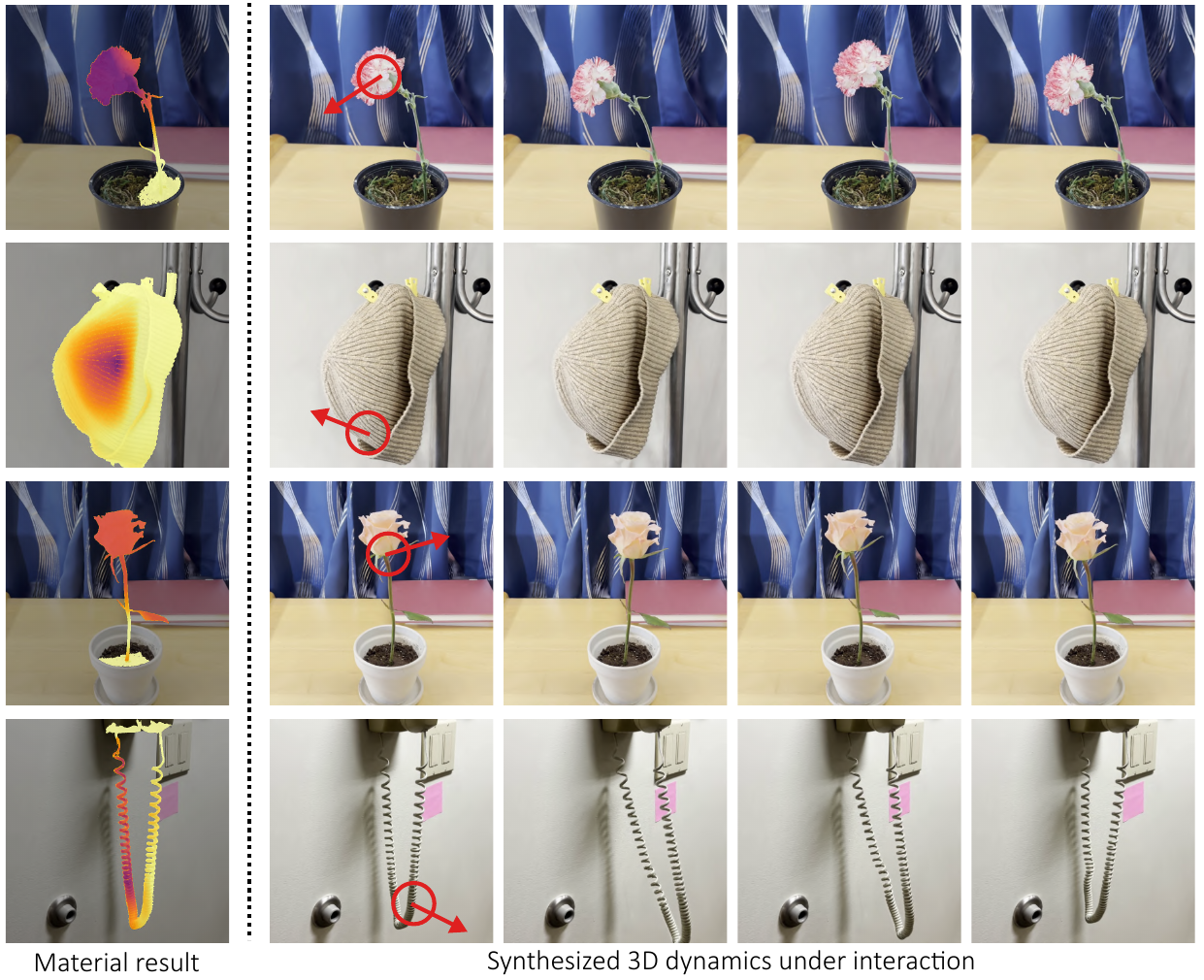}
    % \put(-348,1){ Material result}
    % \put(-190,1){ Synthesized motions}
    \caption{Interactive 3D dynamics synthesis.  \textbf{(Left)} Visualization of the material fields. Brighter color indicates higher Young's modulus within each example.  \textbf{(Right)} We apply an external force (red arrow) on each object, and the following columns demonstrate the object dynamics rendered at a static viewpoint. 
    }
    \label{fig:result_our}
    \vspace{-8pt}
\end{figure}

\subsection{Implementation details}

\myparagraph{Neural material fields.} 
We represent both material field and initial velocity field using triplanes~\cite{peng2020convolutional} each followed by a three-layer MLP. The triplanes have spatial resolutions of $8^3$ and $24^3$ for the material field and velocity field, respectively. 

\myparagraph{3D Gaussian reconstruction.} Similar to PhysGaussian~\cite{physgaussianxie2023}, we employ anisotropic regularization to reduce skinny artifacts in the reconstruction.
Each reconstructed scene contains $0.5$ to $1.5$ million particles (including foreground and background).

\myparagraph{Simulation details.} For computational efficiency, we segment the background and keep only foreground object particles for simulation. 
In our experiments, the foreground object contains around $50$ to $300$ thousand 3D Gaussian particles.
We then discretize the foreground into a $64^3$ grid. 
The number of driving particles are $10$ to $50$ times fewer than the number of 3D Gaussian particles, determined by maintaining an average of at least eight particles per occupied voxel. 
For accurate motion, we use $768$ sub-steps between successive video frames, corresponding to a duration of $4.34 \times 10^{-5}$ second for each sub-step. 
To address the high memory consumption from large number of steps, we apply simulation state checkpointing and re-computation during gradient back-propagation.
We add Dirichlet boundary conditions for stationary grid cells.
We fill the internal volumes of certain solid objects to enhance simulation realism~\cite{physgaussianxie2023}.
% Note that downsampling is compatible with this step, as most removed points lie on object surfaces.

\myparagraph{Generating reference videos. } We render a 3D object with its background from a viewpoint, and then we use Stable Video Diffusion \cite{svdblattmann2023} to animate this rendered image and generate fourteen video frames. We use a small motion bucket number~\cite{svdblattmann2023} (e.g., 5 or 8) so that the generated video contains mostly object motion and little camera motion. 
We use rendered images for the video generation, so that our approach can also be used for generated scenes. Also, rendering images directly from 3D Gaussians simplifies later optimization.
% While not necessary, in our experiments, we choose a viewpoint from the original 3D Gaussian's training set.  

\begin{table}[t]
    \centering
    % \scriptsize
    \fontsize{7.5pt}{11pt}\selectfont
    \caption{Human study 2AFC results of \model (Ours) over real captured videos and  baseline methods (PhysGaussian \cite{physgaussianxie2023} and DreamGaussian4D \cite{dreamgaussian4d2023}) on {\it Motion Realism} and overall {\it Visual Quality}. ``Rose O'', ``Rose W'', and ``Rose R'' denotes the orange, white, and red roses, respectively.}
    \label{tbl:baselines}
    \begin{tabular}{lccccccccc}
    \toprule
         \textbf{Motion realism} & Alocasia & Carnation & Hat & Rose O & Rose W & Rose R & Cord & Tulip & \textbf{Avg.}  \\
      \midrule
      Ours over Real capture & 86\%&61\%&55\%&63\%&47\%&-&29\%&35\%&\textbf{53.7\%}   \\
         \midrule
      Ours over PhysGaussian &  96\%&89\%&57\%&91\%&93\%&73\%&61\%&86\%&\textbf{80.8\%}  \\
      Ours over DreamGaussian & 75\%&77\%&51\%&78\%&51\%&41\%&71\%&64\%&\textbf{63.5\%}  \\
      \midrule
        \textbf{Visual quality} &   \\
      \midrule
      Ours over Real capture & 36\%&53\%&28\%&40\%& 41\%& - &29\%&34\% &37.3\% \\
         \midrule
      Ours over PhysGaussian & 67\% & 69\% & 50\% & 75\% & 73\% & 58\% & 58\% & 70\% & \textbf{65.0\%}     \\
      Ours over DreamGaussian & 82\%&75\%&74\%&76\%&60\%&47\%&76\%&70\%&\textbf{70.0\%}  \\
    \bottomrule
      % Ours over Single view & 81.0\% & 86.0\%
    \end{tabular}
    % \aftertab
\end{table}

\vspace{-10pt}

% \begin{table}[t]
%     \centering
%     \scriptsize    
%     \caption{Human preference of ours over baseline on motion realism and overall visual quality.}
%     \label{tbl:baselines}
%     \begin{tabular}{l|cc}
%          & Visual quality & Motion realism  \\
%          \midrule
%       Ours over PhysGaussian  & 65.0\% & 80.8\%   \\
%       Ours over DreamGaussian & 70.0\% & 63.5\%  \\
%       \hline
%       Ours over Real capture & 37.3\% & 53.7\%   \\
%       % Ours over Single view & 81.0\% & 86.0\%
%     \end{tabular}
%     % \aftertab
% \end{table}

\subsection{Results}

\begin{figure}[t!]
    \centering
    \includegraphics[width=1.0\columnwidth]{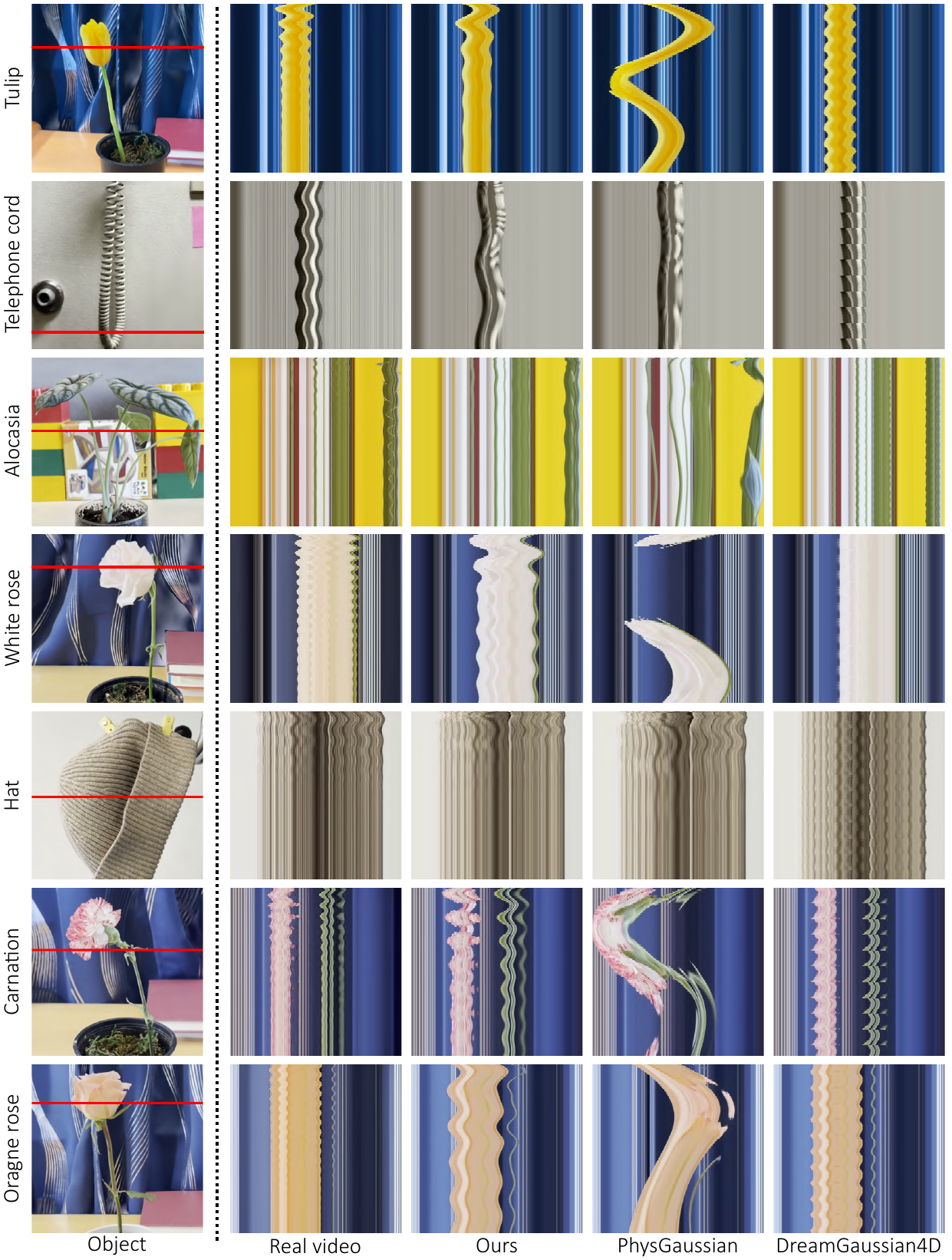}
    % \put(-50,1){How Young's modulus affect the motion}
    % \put(-325,1){\scriptsize Object}
    % \put(-260,1){\scriptsize Real video}
    % \put(-180,1){\scriptsize Ours}
    % \put(-125,1){\scriptsize PhysGaussian }
    % \put(-72,1){\scriptsize DreamGaussian4D }

    \vspace{-2pt}
    
    \caption{We compare our results with real captured videos, PhysGaussian\cite{physgaussianxie2023}, and DreamGaussian4D\cite{dreamgaussian4d2023} using space-time slices. In these slices, the vertical axis represent time, and the horizontal axis shows a spatial slice of the object (denoted by red lines on the ``object'' column).  These slices visualize the magnitude and frequencies of these oscillating motions. Results for our \model(Ours) and PhysGaussian are simulated with the same initial conditions. }
    \label{fig:result_compare}
    \vspace{-8pt}
\end{figure}

We show our qualitative results of the spatially-varying Young's modulus in Fig.~\ref{fig:result_our} (left), and simulated interactive motion in Fig.~\ref{fig:result_our} (right). \emph{Please see our project website videos for a better motion visualization}.
Tab.~\ref{tbl:baselines} presents the user study results in comparison to baseline methods and real captured videos. 

Compared to PhysGaussian, $80.8\%$ of the human participant 2AFC samples prefer \model (ours) in motion realism and $65.0\%$ prefer \model in visual quality. Note that since the static scenes are the same, the visual quality also depends on the generated object motion. 
Fig.~\ref{fig:result_compare} shows temporal slices of the motion patterns. We observe that PhysGaussian produces large, unrealistic slow motion due to the lack of a principled estimation of material properties.

Compared to DreamGaussian4D, $70.0\%$/$63.5\%$ 2AFC samples prefer ours in visual quality and motion realism, respectively. 
From Fig.~\ref{fig:result_compare}, we can observe that DreamGaussian4D generates periodic motion with a constant, small magnitude, while \model can simulate the damping in motion. 
This is because DreamGaussian4D does not simulate the physical dynamics but simply distill a motion sequence from a generative model, so it cannot extrapolate to different motion. 
We further include one more evaluation dimension on ``motion amount'' comparing to DreamGaussian4D, where we ask the participants to judge which video has higher amount of motion, and $73.6\%$ 2AFC samples prefer \model.

% We further extend the user evaluation to compare our simulated motion with real captured videos. For each objects, we recorded four videos illustrating it natural motion after interaction, such as poking or dragging. 
% We then render the simulated motion at the same viewpoint and conduct the user study as previously described.  
Compared to real videos, $53.7\%$ 2AFC samples favored the motion realism of ours results. 
% However, viewers biased with prior knowledge, like the authors, can easily distinguish simulated videos from real videos, since \model produces motion with lower frequencies in most experiments. 
% This low-frequency motion pattern is evident in the space-time slice visualizations at Fig.~\ref{fig:result_compare}. 
% For a more detailed visual comparison, please see the website and supplementary materials.
Interestingly, under “Motion Realism”, 86\% of the users  indicated that the alocasia outputs were more realistic than real captures. This is surprising, as one would expect a 50\% preference if the videos were indistinguishable.  We offer a potential explanation: for thin geometries like alocasia leaves, the Material Point Method tends to produce lower-frequency and slower motions. This can be observed in the video and is evident in the space-time slice visualizations in Fig.~\ref{fig:result_compare}. Humans are poor at judging the naturalness of motion and may be biased towards smoother and slower motions, as shown in prior studies~\cite{stocker2006noise, kobayashi2019perceiving}. 

% One possible reason is that ``Motion Realism" might be too abstract and ambiguous for a user study; Thus, we conducted an additional user study with a more specific prompt: ``Compare the two videos below. One video shows real motion. Please select the real one." The results, shown in supplementary materials, exhibit a similar phenomenon.

% To assess the realism of the simulated motion and the overall visual quality, we conducted a human study to compare our method to the baselines. In addition, we also capture real videos of the objects under interaction for a reference comparison.

% For each object in our experiment, we create a five-second motion sequence, and render them at two static camera viewpoint. 
% We vertically concatenate these two rendered videos.
% In user study, each subject were shown two videos at the same time, one from our method and one from the baseline, and both videos are rendered at the same viewpoint. 
% We asked subjects to rate on the motion realism and overall quality. 
% Specifically, in comparisons involving DreamGaussian4D, an additional criterion, the magnitude of motion, was evaluated. 
% For more details of the user study, please refer to the supplementary materials. 

% We show qualitative comparison with these baseline in Figure~. 

\subsection{Ablation: using multi-view reference videos}

\begin{figure}[t!]
    \centering
    \includegraphics[width=1.0\columnwidth]{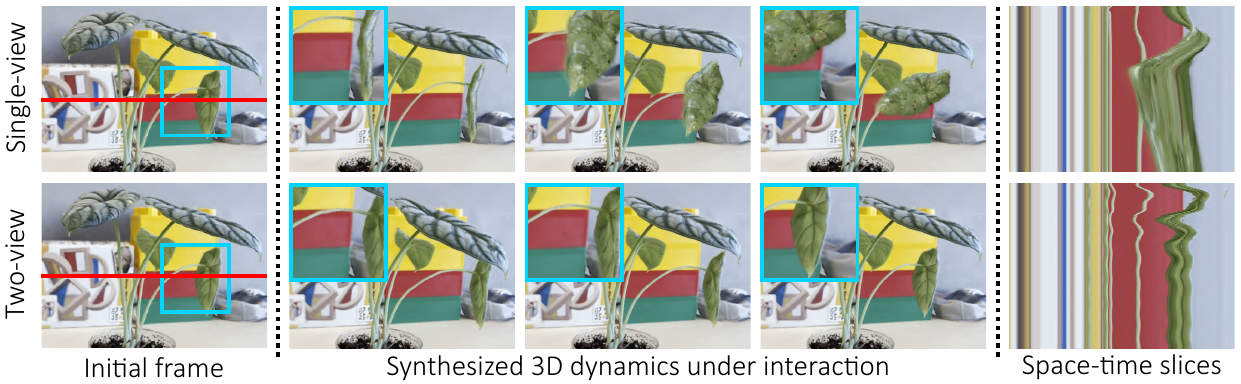}
    % \put(-250,60){Motions from single-view optimization}
    % \put(-250,0){Motions from two-view optimization}
    \vspace{-8pt}
    \caption{Comparison between single-view (top) and two-view (bottom) supervisions. The object (alocasia) exhibits self-occluding structures. We can use generated videos at two views to jointly optimize the material field.  
    % Single-view reference provides less effective supervision signals for materials of occluded parts, as shown in the blue boxes. 
    In the space-time (X-t) slices, the vertical axis represents time, and the horizontal axis shows a spatial slice of the object.}
    \label{fig:result_multiview}
    \vspace{-8pt}
\end{figure}

For objects with self-occlusion, observing salient motion of all object parts from a single video is challenging (e.g., the alocasia scene where a leaf can occlude another leaf).
We may alleviate this problem by rendering from multiple viewpoints to provide comprehensive coverage of the object.
Here, we use multiple videos in the material estimation, jointly optimizing a video-agnostic, spatially-varying Young's modulus for each particle along with video-specific initial velocities.
From the comparison of the alocasia scene in Fig.~\ref{fig:result_multiview}, we can see that using multi-view reference videos (a front view and a back view) helps in such complex self-occluding objects:
\model benefits significantly from having supervision from two views, while using only a single view leads to artifacts. 
In our user study, $81.0\%$ 2AFC samples preferresults with two view supervision in visual quality and $86.0\%$ in motion realism.
 
% We show the results by joint optimizing material parameters from these two videos in Figure~\ref{}.  
% Our user study (Table~\ref{}) confirms the improved realism of motion generated using two viewpoints compared to a single viewpoint in this example.

% \section{Limitations}

% Our approach is not fully automated and requires manual interventions at several stages. It requires selecting a realistic video from several generated ones because Stable Video Diffusion often produces videos with limited object motions, abrupt changes of the objects, or camera shaking. Future advancements in video generation models may mitigate this issue. 
%Our approach requires manually discovering the object to simulate and separate it from the background, and establish boundary conditions for stationary parts, like the pot of flowers. 
%3D object discovery may help for simulatable object extraction. In addition, our approach is computationally demanding. Despite our subsampling strategy, our current algorithm takes approximately one minute on a V100 GPU to produce a single second of video. Further improving efficiency remains an important future problem.
%Finally, in this work we restrict our scope to elastic objects without collisions. Generalizing to more complex scenes is an interesting future direction. 

\section{Conclusion}

% We introduce a novel system that converts static 3D Gaussian representations of photo-realistic elastic objects into interactive models.  By employing differentiable physics and differentiable rendering, our method estimates spatially varying material parameters for the object from video generation models. With the estimated material parameters, our method can simulate realistic physics-based motions of the objects under various physical interactions such as poking and dragging. In experiments with eight everyday elastic objects, our method significantly outperforms prior approaches in terms of perceived motion realism. User studies show that our approach achieves similar human perception scores compared with real captured videos, demonstrating the potential of our approach for creating highly realistic virtual experiences.

In this work, we introduced \model, a novel approach to synthesizing interactive 3D dynamics by endowing static 3D objects with physical material properties. Our method distills the object dynamics priors learned by video generation models to estimate the spatially-varying material properties. 
% The estimated material properties enable realistic physical interaction with the 3D objects. 
We showcased dynamics interaction with a diverse set of elastic objects by \model.
% and a user study, showing that our approach significantly outperforms state-of-the-art methods in terms of motion realism. 
We believe that \model takes a significant step towards creating more engaging and immersive virtual environments, 
% by enabling static 3D objects to respond realistically to novel interactions
opening up a wide range of applications from realistic simulations to interactive virtual experiences.

\myparagraph{Limitations.}
Our approach requires the user to manually specify the object to simulate and separate it from the background, 
and establish boundary conditions for stationary parts, like the pot of flowers. 3D object discovery may help for simulatable object extraction.
In addition, our approach is computationally demanding. Despite our subsampling strategy, our current algorithm takes approximately one minute on a NVIDIA V100 GPU to produce a single second of video. Further improving efficiency remains an important future problem.
Finally, in this work, we restrict our scope to elastic objects without collisions. 
% Generalizing to more complex scenes is an interesting future direction. 

\subsubsection{Acknowledgements.}
This work is in part supported by the NSF PHY-2019786 (The NSF AI Institute for Artificial Intelligence and Fundamental Interactions, \url{http://iaifi.org/}), NSF CIF 1955864 (Occlusion and Directional Resolution in Computational Imaging),  RI \#2211258, \#2338203, ONR MURI N00014-22-1-2740, Quanta Computer, Samsung, and
United States Air Force Artificial Intelligence Accelerator and was
accomplished under Cooperative Agreement Number FA8750-192-1000. We would like to thank Peter Yichen Chen, Zhengqi Li, Pingchuan Ma, Minghao Guo, Ge Yang, and Shai Avidan for help and insightful discussions.

% \clearpage  % TODO REVIEW/FINAL: This \clearpage needs to be removed from both review and camera-ready versions.

\bibliographystyle{splncs04}
\bibliography{main}

\begin{thebibliography}{10}
\providecommand{\url}[1]{\texttt{#1}}
\providecommand{\urlprefix}{URL }
\providecommand{\doi}[1]{https://doi.org/#1}

\bibitem{hyperreelattal2023}
Attal, B., Huang, J.B., Richardt, C., Zollhoefer, M., Kopf, J., O’Toole, M., Kim, C.: Hyperreel: High-fidelity 6-dof video with ray-conditioned sampling. In: Proceedings of the IEEE/CVF Conference on Computer Vision and Pattern Recognition. pp. 16610--16620 (2023)

\bibitem{4dfybahmani2023}
Bahmani, S., Skorokhodov, I., Rong, V., Wetzstein, G., Guibas, L., Wonka, P., Tulyakov, S., Park, J.J., Tagliasacchi, A., Lindell, D.B.: 4d-fy: Text-to-4d generation using hybrid score distillation sampling. arXiv preprint arXiv:2311.17984  (2023)

\bibitem{lumiere2024}
Bar-Tal, O., Chefer, H., Tov, O., Herrmann, C., Paiss, R., Zada, S., Ephrat, A., Hur, J., Li, Y., Michaeli, T., et~al.: Lumiere: A space-time diffusion model for video generation. arXiv preprint arXiv:2401.12945  (2024)

\bibitem{svdblattmann2023}
Blattmann, A., Dockhorn, T., Kulal, S., Mendelevitch, D., Kilian, M., Lorenz, D., Levi, Y., English, Z., Voleti, V., Letts, A., et~al.: Stable video diffusion: Scaling latent video diffusion models to large datasets. arXiv preprint arXiv:2311.15127  (2023)

\bibitem{alignyourlatents2023}
Blattmann, A., Rombach, R., Ling, H., Dockhorn, T., Kim, S.W., Fidler, S., Kreis, K.: Align your latents: High-resolution video synthesis with latent diffusion models. In: Proceedings of the IEEE/CVF Conference on Computer Vision and Pattern Recognition. pp. 22563--22575 (2023)

\bibitem{sora2024}
Brooks, T., Peebles, B., Homes, C., DePue, W., Guo, Y., Jing, L., Schnurr, D., Taylor, J., Luhman, T., Luhman, E., Ng, C., Wang, R., Ramesh, A.: Video generation models as world simulators  (2024), \url{https://openai.com/research/video-generation-models-as-world-simulators}

\bibitem{sax_nerf}
Cai, Y., Wang, J., Yuille, A., Zhou, Z., Wang, A.: Structure-aware sparse-view x-ray 3d reconstruction. In: CVPR (2024)

\bibitem{hexplanecao2023}
Cao, A., Johnson, J.: Hexplane: A fast representation for dynamic scenes. In: Proceedings of the IEEE/CVF Conference on Computer Vision and Pattern Recognition. pp. 130--141 (2023)

\bibitem{i3d2017}
Carreira, J., Zisserman, A.: Quo vadis, action recognition? a new model and the kinetics dataset. In: proceedings of the IEEE Conference on Computer Vision and Pattern Recognition. pp. 6299--6308 (2017)

\bibitem{chan2022efficient}
Chan, E.R., Lin, C.Z., Chan, M.A., Nagano, K., Pan, B., De~Mello, S., Gallo, O., Guibas, L.J., Tremblay, J., Khamis, S., et~al.: Efficient geometry-aware 3d generative adversarial networks. In: Proceedings of the IEEE/CVF Conference on Computer Vision and Pattern Recognition. pp. 16123--16133 (2022)

\bibitem{virtual_elastic_objects_chen2022}
Chen, H.y., Tretschk, E., Stuyck, T., Kadlecek, P., Kavan, L., Vouga, E., Lassner, C.: Virtual elastic objects. In: Proceedings of the IEEE/CVF Conference on Computer Vision and Pattern Recognition. pp. 15827--15837 (2022)

\bibitem{livephotochen2023}
Chen, X., Liu, Z., Chen, M., Feng, Y., Liu, Y., Shen, Y., Zhao, H.: Livephoto: Real image animation with text-guided motion control. arXiv preprint arXiv:2312.02928  (2023)

\bibitem{stochastic_textures_chuang2005}
Chuang, Y.Y., Goldman, D.B., Zheng, K.C., Curless, B., Salesin, D.H., Szeliski, R.: Animating pictures with stochastic motion textures. In: ACM SIGGRAPH 2005 Papers. pp. 853--860 (2005)

\bibitem{volumetriccurless1996}
Curless, B., Levoy, M.: A volumetric method for building complex models from range images. In: Proceedings of the 23rd annual conference on Computer graphics and interactive techniques. pp. 303--312 (1996)

\bibitem{animateanythingdai2023}
Dai, Z., Zhang, Z., Yao, Y., Qiu, B., Zhu, S., Qin, L., Wang, W.: Animateanything: Fine-grained open domain image animation with motion guidance. arXiv e-prints pp. arXiv--2311 (2023)

\bibitem{imagemodaldavis2015}
Davis, A., Chen, J.G., Durand, F.: Image-space modal bases for plausible manipulation of objects in video. ACM Transactions on Graphics (TOG)  \textbf{34}(6), ~1--7 (2015)

\bibitem{visualvibrationdavis2016}
Davis, M.A.: Visual vibration analysis. Ph.D. thesis, Massachusetts Institute of Technology (2016)

\bibitem{4dgsduan2024}
Duan, Y., Wei, F., Dai, Q., He, Y., Chen, W., Chen, B.: 4d gaussian splatting: Towards efficient novel view synthesis for dynamic scenes. arXiv preprint arXiv:2402.03307  (2024)

\bibitem{pienerffeng2023}
Feng, Y., Shang, Y., Li, X., Shao, T., Jiang, C., Yang, Y.: Pie-nerf: Physics-based interactive elastodynamics with nerf. arXiv preprint arXiv:2311.13099  (2023)

\bibitem{kplanefridovich2023}
Fridovich-Keil, S., Meanti, G., Warburg, F.R., Recht, B., Kanazawa, A.: K-planes: Explicit radiance fields in space, time, and appearance. In: Proceedings of the IEEE/CVF Conference on Computer Vision and Pattern Recognition. pp. 12479--12488 (2023)

\bibitem{monoculargao2022}
Gao, H., Li, R., Tulsiani, S., Russell, B., Kanazawa, A.: Monocular dynamic view synthesis: A reality check. Advances in Neural Information Processing Systems  \textbf{35},  33768--33780 (2022)

\bibitem{motionguidancegeng2023}
Geng, D., Owens, A.: Motion guidance: Diffusion-based image editing with differentiable motion estimators. In: The Twelfth International Conference on Learning Representations (2023)

\bibitem{emuvideo2023}
Girdhar, R., Singh, M., Brown, A., Duval, Q., Azadi, S., Rambhatla, S.S., Shah, A., Yin, X., Parikh, D., Misra, I.: Emu video: Factorizing text-to-video generation by explicit image conditioning. arXiv preprint arXiv:2311.10709  (2023)

\bibitem{forwardflowguo2023}
Guo, X., Sun, J., Dai, Y., Chen, G., Ye, X., Tan, X., Ding, E., Zhang, Y., Wang, J.: Forward flow for novel view synthesis of dynamic scenes. In: Proceedings of the IEEE/CVF International Conference on Computer Vision. pp. 16022--16033 (2023)

\bibitem{waltgupta2023}
Gupta, A., Yu, L., Sohn, K., Gu, X., Hahn, M., Fei-Fei, L., Essa, I., Jiang, L., Lezama, J.: Photorealistic video generation with diffusion models. arXiv preprint arXiv:2312.06662  (2023)

\bibitem{fidheusel2017}
Heusel, M., Ramsauer, H., Unterthiner, T., Nessler, B., Hochreiter, S.: Gans trained by a two time-scale update rule converge to a local nash equilibrium. Advances in neural information processing systems  \textbf{30} (2017)

\bibitem{imagenvideo2022}
Ho, J., Chan, W., Saharia, C., Whang, J., Gao, R., Gritsenko, A., Kingma, D.P., Poole, B., Norouzi, M., Fleet, D.J., et~al.: Imagen video: High definition video generation with diffusion models. arXiv preprint arXiv:2210.02303  (2022)

\bibitem{cogvideohong2022}
Hong, W., Ding, M., Zheng, W., Liu, X., Tang, J.: Cogvideo: Large-scale pretraining for text-to-video generation via transformers. arXiv preprint arXiv:2205.15868  (2022)

\bibitem{mlsmpmhu2018}
Hu, Y., Fang, Y., Ge, Z., Qu, Z., Zhu, Y., Pradhana, A., Jiang, C.: A moving least squares material point method with displacement discontinuity and two-way rigid body coupling. ACM Transactions on Graphics (TOG)  \textbf{37}(4),  1--14 (2018)

\bibitem{hu2019taichi}
Hu, Y., Li, T.M., Anderson, L., Ragan-Kelley, J., Durand, F.: Taichi: a language for high-performance computation on spatially sparse data structures. ACM Transactions on Graphics (TOG)  \textbf{38}(6),  1--16 (2019)

\bibitem{taichihu2019}
Hu, Y., Li, T.M., Anderson, L., Ragan-Kelley, J., Durand, F.: Taichi: a language for high-performance computation on spatially sparse data structures. ACM Transactions on Graphics (TOG)  \textbf{38}(6),  1--16 (2019)

\bibitem{scgshuang2023}
Huang, Y.H., Sun, Y.T., Yang, Z., Lyu, X., Cao, Y.P., Qi, X.: Sc-gs: Sparse-controlled gaussian splatting for editable dynamic scenes. arXiv preprint arXiv:2312.14937  (2023)

\bibitem{jiang2017anisotropic}
Jiang, C., Gast, T., Teran, J.: Anisotropic elastoplasticity for cloth, knit and hair frictional contact. ACM Transactions on Graphics (TOG)  \textbf{36}(4),  1--14 (2017)

\bibitem{apicjiang2015}
Jiang, C., Schroeder, C., Selle, A., Teran, J., Stomakhin, A.: The affine particle-in-cell method. ACM Transactions on Graphics (TOG)  \textbf{34}(4),  1--10 (2015)

\bibitem{materialjiang2016}
Jiang, C., Schroeder, C., Teran, J., Stomakhin, A., Selle, A.: The material point method for simulating continuum materials. In: ACM SIGGRAPH 2016 courses. pp. 1--52 (2016)

\bibitem{kinetics2017}
Kay, W., Carreira, J., Simonyan, K., Zhang, B., Hillier, C., Vijayanarasimhan, S., Viola, F., Green, T., Back, T., Natsev, P., et~al.: The kinetics human action video dataset. arXiv preprint arXiv:1705.06950  (2017)

\bibitem{gaussiansplat2023}
Kerbl, B., Kopanas, G., Leimk{\"u}hler, T., Drettakis, G.: 3d gaussian splatting for real-time radiance field rendering. ACM Transactions on Graphics  \textbf{42}(4) (2023)

\bibitem{druckerklar2016}
Kl{\'a}r, G., Gast, T., Pradhana, A., Fu, C., Schroeder, C., Jiang, C., Teran, J.: Drucker-prager elastoplasticity for sand animation. ACM Transactions on Graphics (TOG)  \textbf{35}(4),  1--12 (2016)

\bibitem{kobayashi2019perceiving}
Kobayashi, M., Motoyoshi, I.: Perceiving natural speed in natural movies. i-Perception  \textbf{10}(4),  2041669519860544 (2019)

\bibitem{videopoetkondratyuk2023}
Kondratyuk, D., Yu, L., Gu, X., Lezama, J., Huang, J., Hornung, R., Adam, H., Akbari, H., Alon, Y., Birodkar, V., et~al.: Videopoet: A large language model for zero-shot video generation. arXiv preprint arXiv:2312.14125  (2023)

\bibitem{robustcvdkopf2021}
Kopf, J., Rong, X., Huang, J.B.: Robust consistent video depth estimation. In: Proceedings of the IEEE/CVF Conference on Computer Vision and Pattern Recognition. pp. 1611--1621 (2021)

\bibitem{dynmfkratimenos2023}
Kratimenos, A., Lei, J., Daniilidis, K.: Dynmf: Neural motion factorization for real-time dynamic view synthesis with 3d gaussian splatting. arXiv preprint arXiv:2312.00112  (2023)

\bibitem{le2023differentiable}
Le~Cleac'h, S., Yu, H.X., Guo, M., Howell, T., Gao, R., Wu, J., Manchester, Z., Schwager, M.: Differentiable physics simulation of dynamics-augmented neural objects. IEEE Robotics and Automation Letters  (2023)

\bibitem{globalli2008}
Li, H., Sumner, R.W., Pauly, M.: Global correspondence optimization for non-rigid registration of depth scans. In: Computer graphics forum. vol.~27, pp. 1421--1430. Wiley Online Library (2008)

\bibitem{pacli2023}
Li, X., Qiao, Y.L., Chen, P.Y., Jatavallabhula, K.M., Lin, M., Jiang, C., Gan, C.: Pac-nerf: Physics augmented continuum neural radiance fields for geometry-agnostic system identification. arXiv preprint arXiv:2303.05512  (2023)

\bibitem{nsffli2021}
Li, Z., Niklaus, S., Snavely, N., Wang, O.: Neural scene flow fields for space-time view synthesis of dynamic scenes. In: Proceedings of the IEEE/CVF Conference on Computer Vision and Pattern Recognition. pp. 6498--6508 (2021)

\bibitem{generativeli2023}
Li, Z., Tucker, R., Snavely, N., Holynski, A.: Generative image dynamics. arXiv preprint arXiv:2309.07906  (2023)

\bibitem{dynibarli2023}
Li, Z., Wang, Q., Cole, F., Tucker, R., Snavely, N.: Dynibar: Neural dynamic image-based rendering. In: Proceedings of the IEEE/CVF Conference on Computer Vision and Pattern Recognition. pp. 4273--4284 (2023)

\bibitem{alignling2023}
Ling, H., Kim, S.W., Torralba, A., Fidler, S., Kreis, K.: Align your gaussians: Text-to-4d with dynamic 3d gaussians and composed diffusion models. arXiv preprint arXiv:2312.13763  (2023)

\bibitem{dynamicgaussluiten2023}
Luiten, J., Kopanas, G., Leibe, B., Ramanan, D.: Dynamic 3d gaussians: Tracking by persistent dynamic view synthesis. arXiv preprint arXiv:2308.09713  (2023)

\bibitem{nclawma2023}
Ma, P., Chen, P.Y., Deng, B., Tenenbaum, J.B., Du, T., Gan, C., Matusik, W.: Learning neural constitutive laws from motion observations for generalizable pde dynamics. In: International Conference on Machine Learning. PMLR (2023)

\bibitem{warp2022}
Macklin, M.: Warp: A high-performance python framework for gpu simulation and graphics. \url{https://github.com/nvidia/warp} (March 2022), nVIDIA GPU Technology Conference (GTC)

\bibitem{mildenhall2020nerf}
Mildenhall, B., Srinivasan, P.P., Tancik, M., Barron, J.T., Ramamoorthi, R., Ng, R.: Nerf: Representing scenes as neural radiance fields for view synthesis. In: ECCV (2020)

\bibitem{dynamicfusionnewcombe2015}
Newcombe, R.A., Fox, D., Seitz, S.M.: Dynamicfusion: Reconstruction and tracking of non-rigid scenes in real-time. In: Proceedings of the IEEE conference on computer vision and pattern recognition. pp. 343--352 (2015)

\bibitem{nerfiespark2021}
Park, K., Sinha, U., Barron, J.T., Bouaziz, S., Goldman, D.B., Seitz, S.M., Martin-Brualla, R.: Nerfies: Deformable neural radiance fields. In: Proceedings of the IEEE/CVF International Conference on Computer Vision. pp. 5865--5874 (2021)

\bibitem{hypernerfpark2021}
Park, K., Sinha, U., Hedman, P., Barron, J.T., Bouaziz, S., Goldman, D.B., Martin-Brualla, R., Seitz, S.M.: Hypernerf: A higher-dimensional representation for topologically varying neural radiance fields. arXiv preprint arXiv:2106.13228  (2021)

\bibitem{cleanfidparmar2022}
Parmar, G., Zhang, R., Zhu, J.Y.: On aliased resizing and surprising subtleties in gan evaluation. In: Proceedings of the IEEE/CVF Conference on Computer Vision and Pattern Recognition. pp. 11410--11420 (2022)

\bibitem{peng2020convolutional}
Peng, S., Niemeyer, M., Mescheder, L., Pollefeys, M., Geiger, A.: Convolutional occupancy networks. In: Computer Vision--ECCV 2020: 16th European Conference, Glasgow, UK, August 23--28, 2020, Proceedings, Part III 16. pp. 523--540. Springer (2020)

\bibitem{dreamfusionpoole2022}
Poole, B., Jain, A., Barron, J.T., Mildenhall, B.: Dreamfusion: Text-to-3d using 2d diffusion. In: The Eleventh International Conference on Learning Representations (2022)

\bibitem{dnerfpumarola2021}
Pumarola, A., Corona, E., Pons-Moll, G., Moreno-Noguer, F.: D-nerf: Neural radiance fields for dynamic scenes. In: Proceedings of the IEEE/CVF Conference on Computer Vision and Pattern Recognition. pp. 10318--10327 (2021)

\bibitem{materialram2015}
Ram, D., Gast, T., Jiang, C., Schroeder, C., Stomakhin, A., Teran, J., Kavehpour, P.: A material point method for viscoelastic fluids, foams and sponges. In: Proceedings of the 14th ACM SIGGRAPH/Eurographics Symposium on Computer Animation. pp. 157--163 (2015)

\bibitem{dreamgaussian4d2023}
Ren, J., Pan, L., Tang, J., Zhang, C., Cao, A., Zeng, G., Liu, Z.: Dreamgaussian4d: Generative 4d gaussian splatting. arXiv preprint arXiv:2312.17142  (2023)

\bibitem{makeavideo2022}
Singer, U., Polyak, A., Hayes, T., Yin, X., An, J., Zhang, S., Hu, Q., Yang, H., Ashual, O., Gafni, O., et~al.: Make-a-video: Text-to-video generation without text-video data. arXiv preprint arXiv:2209.14792  (2022)

\bibitem{text4dsinger2023}
Singer, U., Sheynin, S., Polyak, A., Ashual, O., Makarov, I., Kokkinos, F., Goyal, N., Vedaldi, A., Parikh, D., Johnson, J., et~al.: Text-to-4d dynamic scene generation. arXiv preprint arXiv:2301.11280  (2023)

\bibitem{stocker2006noise}
Stocker, A.A., Simoncelli, E.P.: Noise characteristics and prior expectations in human visual speed perception. Nature neuroscience  \textbf{9}(4),  578--585 (2006)

\bibitem{materialstomakhin2013}
Stomakhin, A., Schroeder, C., Chai, L., Teran, J., Selle, A.: A material point method for snow simulation. ACM Transactions on Graphics (TOG)  \textbf{32}(4),  1--10 (2013)

\bibitem{tang2023dreamgaussian}
Tang, J., Ren, J., Zhou, H., Liu, Z., Zeng, G.: Dreamgaussian: Generative gaussian splatting for efficient 3d content creation. arXiv preprint arXiv:2309.16653  (2023)

\bibitem{fvdunterthiner2018}
Unterthiner, T., Van~Steenkiste, S., Kurach, K., Marinier, R., Michalski, M., Gelly, S.: Towards accurate generative models of video: A new metric \& challenges. arXiv preprint arXiv:1812.01717  (2018)

\bibitem{phenakivillegas2022}
Villegas, R., Babaeizadeh, M., Kindermans, P.J., Moraldo, H., Zhang, H., Saffar, M.T., Castro, S., Kunze, J., Erhan, D.: Phenaki: Variable length video generation from open domain textual descriptions. In: International Conference on Learning Representations (2022)

\bibitem{flowwang2023}
Wang, C., MacDonald, L.E., Jeni, L.A., Lucey, S.: Flow supervision for deformable nerf. In: Proceedings of the IEEE/CVF Conference on Computer Vision and Pattern Recognition. pp. 21128--21137 (2023)

\bibitem{diffusionwang2024}
Wang, C., Zhuang, P., Siarohin, A., Cao, J., Qian, G., Lee, H.Y., Tulyakov, S.: Diffusion priors for dynamic view synthesis from monocular videos. arXiv preprint arXiv:2401.05583  (2024)

\bibitem{nuwawu2022}
Wu, C., Liang, J., Ji, L., Yang, F., Fang, Y., Jiang, D., Duan, N.: N{\"u}wa: Visual synthesis pre-training for neural visual world creation. In: European conference on computer vision. pp. 720--736. Springer (2022)

\bibitem{spacexian2021}
Xian, W., Huang, J.B., Kopf, J., Kim, C.: Space-time neural irradiance fields for free-viewpoint video. In: Proceedings of the IEEE/CVF Conference on Computer Vision and Pattern Recognition. pp. 9421--9431 (2021)

\bibitem{physgaussianxie2023}
Xie, T., Zong, Z., Qiu, Y., Li, X., Feng, Y., Yang, Y., Jiang, C.: Physgaussian: Physics-integrated 3d gaussians for generative dynamics. arXiv preprint arXiv:2311.12198  (2023)

\bibitem{xie2022neural}
Xie, Y., Takikawa, T., Saito, S., Litany, O., Yan, S., Khan, N., Tombari, F., Tompkin, J., Sitzmann, V., Sridhar, S.: Neural fields in visual computing and beyond. In: Computer Graphics Forum. vol.~41, pp. 641--676. Wiley Online Library (2022)

\bibitem{deformablegaussianyang2023}
Yang, Z., Gao, X., Zhou, W., Jiao, S., Zhang, Y., Jin, X.: Deformable 3d gaussians for high-fidelity monocular dynamic scene reconstruction. arXiv preprint arXiv:2309.13101  (2023)

\bibitem{4dgenyin2023}
Yin, Y., Xu, D., Wang, Z., Zhao, Y., Wei, Y.: 4dgen: Grounded 4d content generation with spatial-temporal consistency. arXiv preprint arXiv:2312.17225  (2023)

\bibitem{cogsyu2023}
Yu, H., Julin, J., Milacski, Z.{\'A}., Niinuma, K., Jeni, L.A.: Cogs: Controllable gaussian splatting. arXiv preprint arXiv:2312.05664  (2023)

\bibitem{dylinyu2023}
Yu, H., Julin, J., Milacski, Z.A., Niinuma, K., Jeni, L.A.: Dylin: Making light field networks dynamic. In: Proceedings of the IEEE/CVF Conference on Computer Vision and Pattern Recognition. pp. 12397--12406 (2023)

\bibitem{i2vgenzhang2023}
Zhang, S., Wang, J., Zhang, Y., Zhao, K., Yuan, H., Qin, Z., Wang, X., Zhao, D., Zhou, J.: I2vgen-xl: High-quality image-to-video synthesis via cascaded diffusion models. arXiv preprint arXiv:2311.04145  (2023)

\bibitem{consistentzhang2021}
Zhang, Z., Cole, F., Tucker, R., Freeman, W.T., Dekel, T.: Consistent depth of moving objects in video. ACM Transactions on Graphics (TOG)  \textbf{40}(4),  1--12 (2021)

\end{thebibliography}

\clearpage

\appendix
\section*{Appendix}
\addcontentsline{toc}{section}{Appendix}

\section{Metrics}\label{sec:metrics}

We  compare the visual quality of our method with two baseline methods, PhysGaussian \cite{physgaussianxie2023} and DreamGaussian4D \cite{dreamgaussian4d2023}, by computing the Frechet Video Distance (FVD) \cite{fvdunterthiner2018} against real captured videos. We compute the FVD with a 16-frame window, 2-frame stride, based on the I3D \cite{i3d2017} model trained on the Human Kinetics Dataset \cite{kinetics2017}. All videos are resized (short edge to $144$ pixels) and center-cropped to $128 \times 128$ pixels prior to FVD computation. We compare each method against real captured videos, creating 272 clips per scene for evaluation. The results are shown in Table~\ref{tbl:fvd}. 

We further compare methods using the Frechet Inception Distance (FID) \cite{cleanfidparmar2022, fidheusel2017}, as shown in Table~\ref{tbl:fid}. FID calculation incorporates all frames across all objects, totaling 4200 frames per method.

\begin{table}[h]
    \centering
    % \scriptsize
    \fontsize{7.5pt}{11pt}\selectfont
    \caption{Frechet Video Distance (FVD) between real captured video and \model (Ours) and  baseline methods (PhysGaussian \cite{physgaussianxie2023} and DreamGaussian4D \cite{dreamgaussian4d2023})}
    \label{tbl:fvd}
    \begin{tabular}{lcccccccc}
    \toprule
         \textbf{FVD ($\downarrow$)} & Alocasia & Carnation & Hat & Rose O. & Rose W. &  Cord & Tulip & \textbf{Avg.}  \\
      \midrule
      Ours & 272 & 282 & 54 & 231 & 640  & 185 & 228 &\textbf{270.3}   \\
         \midrule
      PhysGaussian &  560 & 629 & 50 & 408 & 961 & 
184 & 586 & 482.6  \\
    DreamGaussian & 308 & 359 & 75 & 200 & 1379 & 210 & 497 &432.6  \\
    \bottomrule
      % Ours over Single view & 81.0\% & 86.0\%
    \end{tabular}
    % \aftertab
\end{table}

\begin{table}[h]
    \centering
    % \scriptsize
    \fontsize{7.5pt}{11pt}\selectfont
    \caption{Frechet Inception Distance (FID) between real captured video and \model (Ours) and  baseline methods (PhysGaussian \cite{physgaussianxie2023} and DreamGaussian4D \cite{dreamgaussian4d2023})}
    \label{tbl:fid}
    \begin{tabular}{lc}
    \toprule
         Method & \textbf{FID ($\downarrow$)} \\
      \midrule
      Ours & \textbf{47.7}   \\
         \midrule
      PhysGaussian &  63.2   \\
    DreamGaussian & 52.8   \\
    \bottomrule
      % Ours over Single view & 81.0\% & 86.0\%
    \end{tabular}
    % \aftertab
\end{table}

\section{User Study}\label{sec:human_study}
We use Prolific\footnote{\url{https://www.prolific.com/}} to recruit participants for the human preference evaluation. We use Google forms to present the survey. The survey is fully anonymized for both the participants and the host. We attach an example anonymous survey link in the
footnote\footnote{An example user study survey (comparing to PhysGaussian): \url{https://forms.gle/CZfwxGHX2LaA7KxGA}. Google forms require signing in to participate, but it does not record any participant's identity.} for reference. Reviewer can enter any text such as ``test'' for Prolific ID.

% \section{Pseudo code of the algorithm}

% MPM, Interpolation, Gradient Checkpoints. 

% \section{Website}
% We encourage the readers to explore videos in the attached website. Open the \url{index.html} to see the website. 

\section{Algorithm details}\label{sec:algorithm}
We present python-style pseudo-code for accelerating material point methods with K-Means downsampling in Algorithm~\ref{alg:kmeans_mpm}.

\begin{algorithm}[h]
\caption{Acclerate material point method with downsampling}
\label{alg:kmeans_mpm}
\definecolor{codeblue}{rgb}{0.25,0.5,0.5}
\lstset{
  backgroundcolor=\color{white},
  basicstyle=\fontsize{9pt}{11pt}\ttfamily\selectfont,
  columns=fullflexible,
  breaklines=true,
  captionpos=b,
  commentstyle=\fontsize{9pt}{11pt}\color{codeblue},
  keywordstyle=\fontsize{7.2pt}{7.2pt},
}
\begin{lstlisting}[language=python]
# x, alpha, R, Sigma, c: the position, opacity, rotation, covariance and color of each Gaussian particle.  x of shape [N, 3]
# num_drive_pts: int, top_k: int default as 8

clusters = KMeans(x, num_drive_pts)
drive_x = clusters.x  # [M, 3]

# pre-compute the index of neighboor points
cdist = -1.0 * torch.cdist(x, drive_x) # [N, M]
_, top_k_index = torch.topk(cdist, top_k, -1)

# query initial velocity and material params, and simulate
drive_v = VeloField(drive_x)
drive_material = MaterialField(drive_x)
drive_x_simulated = Simulate(drive_x, drive_v, drive_material)

neighboor_drive_x = drive_x[top_k_index] # [N, top_k, 3]
neighboor_drive_x_simulated = drive_x_simulated[top_k_index]
# R: [N, 3, 3], t: [N, 3]
R_sim, t_sim = fitRigidTransform(drive_x, drive_x_simulated) 

# apply transform to interpolate points
x = x + t_sim
R = R_sim @ R
# render
frame = Render(x, alpha, R @ Sigma @ R.T, c)







\end{lstlisting}
\end{algorithm}

\clearpage  % TODO REVIEW/FINAL: This \clearpage needs to be removed from both review and camera-ready versions.


\end{document}